%% file: paper.tex
\documentclass[]{research}
\usepackage{amsmath}
\usepackage{amsfonts}
\usepackage{hyperref}

\usepackage{url}
\usepackage{booktabs}
\usepackage{multirow}
\usepackage{pgfplots}
\usepackage{graphicx}
\usepackage{array}
\pgfplotsset{compat=1.18}
\usepackage{arydshln}
\usepackage{xspace}
\usepackage{wrapfig}
\usepackage{algorithm}
\usepackage{algpseudocode} 
\usepackage{siunitx}

\crefname{figure}{Figure}{Figures}
\crefname{table}{Table}{Tables}
\crefname{section}{Section}{Sections}
\crefname{algorithm}{Algorithm}{Algorithms}
\crefname{appendix}{Appendix}{Appendices}

\makeatletter
\def\adl@drawiv#1#2#3{%
        \hskip.5\tabcolsep
        \xleaders#3{#2.5\@tempdimb #1{1}#2.5\@tempdimb}%
                #2\z@ plus1fil minus1fil\relax
        \hskip.5\tabcolsep}
\newcommand{\cdashlinelr}[1]{%
  \noalign{\vskip\aboverulesep
           \global\let\@dashdrawstore\adl@draw
           \global\let\adl@draw\adl@drawiv}
  \cdashline{#1}
  \noalign{\global\let\adl@draw\@dashdrawstore
           \vskip\belowrulesep}}
\makeatother

\usepackage{listings}
\usepackage{xcolor}
\lstdefinestyle{icmlcode}{
  basicstyle=\ttfamily\small,
  columns=fullflexible,
  breaklines=true,
  frame=single,
  rulecolor=\color{black!20},
  frameround=tttt,
  showstringspaces=false,
  tabsize=2,
  keywordstyle=\color{blue!60!black},
  commentstyle=\color{green!40!black},
  stringstyle=\color{orange!60!black},
}

\lstdefinestyle{icmlprompt}{
  basicstyle=\ttfamily\normalsize,
  columns=fullflexible,
  breaklines=true,
  frame=single,
  rulecolor=\color{black!20},
  frameround=tttt,
  showstringspaces=false,
  tabsize=2,
  numbers=left,
  numberstyle=\tiny\color{black!50},
}

\lstdefinelanguage{CUDA}{
  language=C++,
  morekeywords={__global__,__host__,__device__,__shared__,__align__,__launch_bounds__,
                __syncthreads,__shfl_xor_sync,__nv_bfloat16,wmma,fragment,__pipeline_memcpy_async,
                __pipeline_commit,__pipeline_wait_prior},
}

\newcommand{\system}{\textsc{K-Search}\xspace}

\title{\system: LLM Kernel Generation via Co-Evolving Intrinsic World Model}
\author[]{Shiyi Cao}
\author[]{Ziming Mao}
\author[]{Joseph E. Gonzalez}
\author[]{Ion Stoica}

\affiliation[]{UC Berkeley}

\abstract{Optimizing GPU kernels is critical for efficient modern machine learning systems yet remains challenging due to the complex interplay of design factors and rapid hardware evolution. Existing automated approaches typically treat Large Language Models (LLMs) merely as stochastic code generators within heuristic-guided evolutionary loops. These methods often struggle with complex kernels requiring coordinated, multi-step structural transformations, as they lack explicit planning capabilities and frequently discard promising strategies due to inefficient or incorrect intermediate implementations. To address this, we propose \emph{Search via Co-Evolving World Model} and build \system based on this method. By replacing static search heuristics with a co-evolving world model, our framework leverages LLMs' prior domain knowledge to guide the search, actively exploring the optimization space. This approach explicitly decouples high-level algorithmic planning from low-level program instantiation, enabling the system to navigate non-monotonic optimization paths while remaining resilient to temporary implementation defects. We evaluate \system on diverse, complex kernels from FlashInfer, including GQA, MLA, and MoE kernels. Our results show that \system significantly outperforms state-of-the-art evolutionary search methods, achieving an average 2.10$\times$ improvement and up to a 14.3$\times$ gain on complex MoE kernels. On the GPUMode \href{https://www.gpumode.com/leaderboard/496?tab=rankings}{TriMul} task, \system achieves state-of-the-art performance on H100, reaching \SI{1030}{\micro\second} and surpassing both prior evolution and human-designed solutions.}

\metadata[Code]{\href{https://github.com/caoshiyi/K-Search}{https://github.com/caoshiyi/K-Search}}

\begin{document}

\maketitle

\input{latex/intro-v1}

\input{latex/related}

\input{latex/method-v4}

\input{latex/exp}
\input{latex/limitations}
\input{latex/acknowledgement}

\bibliographystyle{plainnat}
\bibliography{paper}

\clearpage
\newpage
\beginappendix
\input{latex/appendix}

\end{document}

%% file: latex/intro-v1.tex
\section{Introduction}

High-performance GPU kernels are fundamental to modern machine learning systems for training and serving large models, as reflected in widely used optimized libraries such as FlashInfer~\cite{ye2025flashinfer} for high-throughput LLM serving~\cite{kwon2023efficient,zheng2024sglang}, FlashAttention~\cite{dao2022flashattention,dao2023flashattention,shah2024flashattention} for memory-efficient attention, and FlashLinearAttention~\cite{fla2024flashlinear} for emerging model architecture variants~\cite{gu2024mamba,peng2023rwkv}.

Unfortunately, GPU kernels are notoriously challenging to manually optimize and tune. First, achieving near-peak performance on modern GPUs requires navigating a large design space of tiling, memory layout, synchronization, and architecture-specific primitives~\cite{nvidia_ptx_isa}. Second, fast hardware evolution adds additional optimization complexity; new architectures (e.g., transitioning from NVIDIA Hopper to Blackwell) introduce new instructions and architectural characteristics that fundamentally alter performance trade-offs, rendering previously optimized kernels sub-optimal. Third, testing a kernel optimization might require significant implementation effort with many manual trials and errors~\cite{ouyang2025kernelbench}. Compiling and profiling generated kernels is also computationally expensive, normally mandating strictly limited testing budgets~\cite{zheng2020ansor}. Therefore, automated kernel generation methods that can adapt efficiently to new workloads and hardware with low search budgets become increasingly important.

Existing LLM evolutionary approaches, such as  OpenEvolve~\cite{openevolve2025overview}, couple language models with genetic algorithms or quality-diversity search. These methods typically treat LLMs purely as stochastic code generators, relying on heuristic mechanisms such as MAP-Elites~\cite{mouret2015illuminating} to select and mutate candidates directly in the program space. However, high-performance kernels often require coordinated, multi-step structural transformations, such as refactoring memory layout \emph{before} applying vectorization, where intermediate steps may not yield immediate performance gains. By treating the LLM merely as a code generator without an explicit planning mechanism, existing evolutionary methods typically cannot plan multi-step optimization sequences in which intermediate edits fail to improve the objective, and they often prematurely discard theoretically sound strategies due to temporary compilation errors, limiting their ability to discover deep structural optimizations necessary for achieving state-of-the-art performance.

To address this, we propose \textbf{Search via Co-Evolving World Model} and build the \system framework, inspired by recent works~\cite{fang2025webevolver,hao2023reasoning} on using large language models as world models to guide planning and decision making. We formulate kernel generation as a planning problem over a structured search tree, governed by a World Model instantiated from an LLM to leverage its prior domain knowledge for efficient search. In this setup, the World Model is responsible for maintaining the search frontier and estimating the priority scores of high-level optimization intents. Crucially, this model is \emph{co-evolving} with the search process: it continuously refines its transition dynamics by assimilating execution feedback via in-context learning, allowing it to dynamically update its prior belief and calibrate the search strategy. This approach explicitly decouples high-level planning from low-level program instantiation, enabling the system to navigate complex, non-monotonic optimization paths while remaining resilient to temporary implementation defects.

We evaluate \system on a diverse set of complex workloads from FlashInfer~\cite{ye2025flashinfer}, including GQA~\cite{yang2025qwen3}, MLA~\cite{flashinfer_deepwiki_mla}, and MoE~\cite{liu2024deepseek} kernels. Our results demonstrate that \system significantly outperforms state-of-the-art baselines, achieving an average \textbf{2.10$\times$ improvement} over OpenEvolve and an average \textbf{2.21$\times$ improvement} over ShinkaEvolve. Notably, on the challenging MoE kernel, \system delivers a \textbf{14.3$\times$ improvement} over OpenEvolve. We also test \system on the GPUMode \href{https://www.gpumode.com/leaderboard/496?tab=rankings}{TriMul} task, and it achieves state-of-the-art performance (\SI{1030}{\micro\second} on H100), surpassing both prior automated and human-designed solutions. Collectively, these results validate the efficacy of disentangling high-level intent from low-level implementation to enable deep structural optimization.

%% file: latex/related.tex
\section{Related Work}
\label{sec:related_work}

\paragraph{Fast iteration and specialized kernel libraries.}
Recent years have seen substantial engineering effort devoted to highly optimized, workload-specific GPU kernel libraries. Examples include FlashAttention~\cite{dao2022flashattention,dao2023flashattention,shah2024flashattention} for dense attention, FlashLinearAttention~\cite{fla2024flashlinear} for a wide range of linear and state-space attention variants such as Mamba~\cite{gu2024mamba} and RWKV~\cite{peng2023rwkv}, and FlashInfer~\cite{ye2025flashinfer} for high-throughput LLM serving with paged KV-cache and dynamic batching. Although each library targets a relatively narrow class of operators, achieving near-peak performance requires careful architecture-specific tuning. As model architectures continue to diversify and introduce new attention mechanisms and custom sequence operators, manually developing and maintaining specialized kernels becomes increasingly costly, motivating automated  LLM-based kernel generation methods that can rapidly adapt to new workloads and hardware.

\paragraph{Compiler autotuning, DSLs, and kernel optimization.}
Automated kernel optimization has a long history in compilers and high-performance systems. TVM~\cite{chen2018tvm} and its associated auto-schedulers, such as Ansor~\cite{zheng2020ansor}, optimize tensor programs by employing learned cost models to search large scheduling spaces.
Parallel to these search-based approaches, domain-specific frameworks like Triton~\cite{tillet2019triton} and layout abstractions like CuTe~\cite{thakkar2023cute} enable higher-level expression of GPU kernels, allowing developers to explicitly manage architecture-aware tiling and memory hierarchies.  
Collectively, these systems illustrate the immense complexity of GPU kernel optimization, highlighting the necessity for methods capable of coordinating tiling, memory layouts, and specialized instructions to maximize performance.

\paragraph{LLMs for GPU kernel generation.}
Most existing LLM-based GPU kernel generation systems~\cite{wei2025astra,zhang2025cudaforge,lange2025towards,liao2025kernelevolve,li2025tritonforge} employ simple iterative search or refinement pipelines, where LLMs generate kernel variants based on compilation results, execution, and profiling feedback. Recent work augments this paradigm with evolutionary strategies to improve exploration, such as EvoEngineer~\cite{guo2025evoengineer}, which maintains and manages a population of candidate kernels during search. In parallel, several works~\cite{li2025cuda,li2025autotriton,baronio2025kevin} leverage reinforcement learning to train models to generate optimized CUDA or Triton kernels, primarily focusing on enhancing the model's one-shot generation or local refinement capabilities.

\paragraph{LLM-guided evolutionary and population-based program search.}
Recent work has explored coupling large language models with execution-based evaluation and evolutionary search to discover high-quality programs. FunSearch~\citep{romera2024mathematical} pairs an LLM with an evaluator in an evolutionary loop to improve solutions in mathematical and combinatorial domains. AlphaEvolve~\cite{novikov2025alphaevolve} generalizes this paradigm to codebase evolution using LLM-generated edits and a program database that seeds subsequent generations. OpenEvolve~\cite{openevolve2025overview}, the open-source realization of these ideas, instantiates archive-based evolution with explicit island models and quality-diversity mechanisms such as MAP-Elites~\cite{mouret2015illuminating}. ShinkaEvolve~\cite{lange2025shinkaevolve} proposes a population-based evolutionary framework that combines performance-driven selection with novelty-aware rejection to improve sample efficiency. Critically, these methods fundamentally treat the LLM merely as a stochastic code generator. They search directly in the space of program implementations, relying on evolutionary heuristics to drive progress rather than leveraging the LLM's capacity for high-level planning or reasoning. 

\paragraph{Large Language Models as World Models}
Recent work suggests that LLMs can function as implicit or explicit world models for planning and decision making. RAP~\cite{hao2023reasoning} frames reasoning as planning with an LLM world model. Complementary approaches externalize dynamics by inducing structured domain models (e.g., PDDL) from language and refining them for classical planning \cite{guan2023leveraging}. Recent works~\cite{fang2025webevolver,gu2024your} further demonstrate that LLMs can act as world models to guide agentic planning tasks by simulating action outcomes and evaluating candidate trajectories. Together, these studies highlight an emerging perspective of treating LLMs as structured world models that support search and model-based planning.

%% file: latex/method-v4.tex
\section{\system}
\label{sec:method}

\subsection{Problem Setup}
\label{sec:problem_setup}

We formulate GPU kernel synthesis as an optimization problem under a fixed evaluation budget. We list the core notations in \cref{tab:notation}.

\begin{wraptable}{r}{0.54\linewidth}
\vspace{-10pt}
\centering
\caption{Notation and Definitions}
\label{tab:notation}
\begin{tabular}{l l}
\toprule
\textbf{Symbol} & \textbf{Definition} \\
\midrule
$x \in \mathcal{X}$ & Kernel program implementation \\
$o \in \mathcal{O}$ & Observation tuple from execution \\
$\mathcal{E}: \mathcal{X} \to \mathcal{O}$ & Evaluator function \\
$J: \mathcal{X} \to \mathbb{R}$ & Scalar objective score \\
$\mathcal{H}_t$ & History $\{(x_i, o_i)\}_{i=1}^t$ \\
$S_t$ & Search state \\
$\mathcal{A}(S_t)$ & Frontier of actions \\
$V(a \mid S_t) \in [0,1]$ & Model-estimated priority score \\
\bottomrule
\end{tabular}
\vspace{-10pt}
\end{wraptable}

For a given kernel program $x$, the evaluator returns an observation tuple:
\[
o \;=\; (s, p, m) \;=\; \mathcal{E}(x),
\]
where $s \in \{0, 1\}$ indicates correctness, $p \in \mathbb{R}^+$ is the performance metric (i.e., latency), and $m$ contains metadata (e.g., compiler logs, profiler output). We define the maximization objective $J(x)$ as the speedup relative to a reference SoTA baseline ($p_{\text{ref}}$):
\[
J(x) \;=\; s \cdot \frac{p_{\text{ref}}}{p} \cdot 100.
\]
The optimization goal is to identify a program $x^\star = \operatorname*{arg\,max}_{x \in \mathcal{X}} J(x)$ utilizing a fixed budget of $B$ evaluations.

\subsection{Search via Co-Evolving World Model}
\label{sec:modeling}

\begin{figure*}[ht!] 
\centering 
\includegraphics[width=0.87\linewidth]{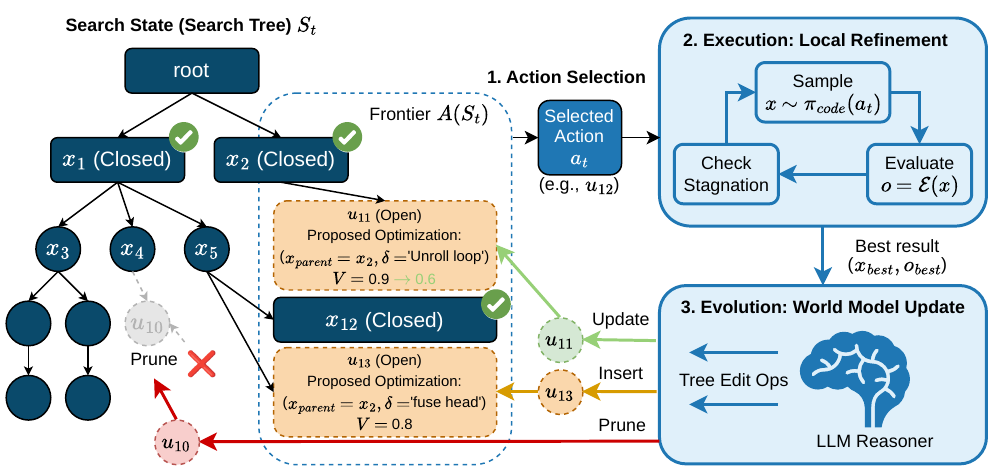} 
\caption{\textbf{Overview of \system.} The framework operates on a \textbf{Search State} $S_t$ structured as a search tree. The tree consists of \textsc{Closed} nodes (blue, visited states with attached program like $x_{12}$) and a Frontier of \textsc{Open} nodes (orange, pending hypotheses like $u_{13}$). The workflow iterates through three phases: (1) \textbf{Action Selection}, where the most promising action node is retrieved from the frontier based on world model estimated priority score $V$; (2) \textbf{Local Refinement}, where a stochastic policy $\pi_{\text{code}}$ samples concrete implementations until stagnation; and (3) \textbf{World Model Update}, where the LLM reasons over the trajectory to update the search tree via \textit{Insert} (adding new actions), \textit{Update} (adjusting $V$, e.g., $u_{11}$ dropping from 0.9 to 0.6), and \textit{Prune} (removing less promising nodes like $u_{10}$).} 
\label{fig:system_overview} 
\end{figure*}

Large Language Models possess rich intrinsic prior knowledge regarding optimization heuristics and strong planning capabilities~\cite{zhao2023large,bohnet2025enhancing}. However, existing evolution methods often underutilize these capabilities by treating the LLM merely as a code generator. We show that effective search requires disentangling \emph{algorithmic planning} (leveraging intrinsic and evolving understanding) from \emph{implementation}.

\paragraph{Baseline: Heuristic Search in Program Space.}
Existing approaches search directly in the space of the program. At step $t$, they select a context subset $C_t \subseteq \mathcal{H}_t$ using evolutionary heuristics such as MAP-Elites~\cite{mouret2015illuminating,romera2024mathematical,openevolve2025overview}) and generate a new candidate by conditioning on the raw text of previous programs and their associated execution feedback:
\[
x_{t+1} \sim \pi_{\text{LLM}}\left(x \;\middle|\; \left\{ (x_k, o_k) \right\}_{(x_k, o_k) \in C_t} \right).
\]
Here, the observation $o_k$ is serialized into a textual prompt (e.g., compiler error messages or profiler logs). This formulation inherently couples high-level optimization intent with low-level implementation. Lacking an explicit planning mechanism, a theoretically sound strategy may be discarded simply because of a transient syntax error in $x_{t+1}$.

\paragraph{Ours: Search via Co-Evolving World Model.}
We structure code generation as a search process guided by an LLM repurposed as a world model with an intrinsic understanding of the design space. The \emph{World Model} estimates the next state of the search process after applying an action to the current
state~\cite{ha2018world,hao2023reasoning,matsuo2022deep}.

Formally, the LLM world model represents a search state ($S_t$) transition distribution $P_{\text{model}}(S_{t+1} \mid S_t, a_t)$.
A search state $S_t$ encapsulates the current snapshot of the model's understanding on the search process, including the history of explored actions and their performance, the \emph{frontier} $\mathcal{A}(S_t)$ actions (i.e., the set of pending, unexplored actions available), and the estimated priority score $V$ over the frontier actions. Crucially, this model is \emph{co-evolving} with the search process: it continuously refines its understanding and beliefs based on past trials and accumulated experience via in-context learning.

The search proceeds through three iterative phases:

\textbf{1. Action Selection.} 
The hypothesis with the highest priority score $V$ from the search state's current frontier $\mathcal{A}(S_t)$ is selected as the next action:
\[
a_{t} = \operatorname*{arg\,max}_{a \in \mathcal{A}(S_t)} V(a \mid S_t).
\]
Here, an action $a_t = (x_{\text{parent}}, \delta)$ represents a specific intent $\delta$ (e.g., ``Resolve bank conflicts via padding'') applied to a specific parent program $x_{\text{parent}}$. $V$ is a scalar priority score predicted by the world model when the action is created (or updated). It represents the model's intrinsic assessment of the potential of the actions.

\textbf{2. Program Instantiation.} 
A concrete program is then synthesized by applying the selected plan to the parent implementation found in $S_t$:
\[
x_{t} \sim \pi_{\text{code}}(x \mid a_t), \quad o_t = \mathcal{E}(x_t).
\]

\textbf{3. World Model Co-Evolution.} 
Upon observing the result $o_t$ of $x_t$ (i.e., the outcome of action $a_t$), the world model then performs the search state transition to obtain $S_{t+1}$ conditioned on the newly accumulated experience, which updates $\mathcal{A}$ and calibrates $V$:
\[
S_{t+1} \sim P_{\text{model}}(S \mid S_t, a_t; x_t, o_t).
\]

This co-evolution allows the world model to update its prior assumptions and beliefs effectively, and perform the state transition to progressively sharpen the search policy.

\subsection{System Design}
\label{sec:system}

\begin{wrapfigure}{r}{0.55\columnwidth}
\vspace{-6pt}
\begin{minipage}{0.55\columnwidth}
\small
\begin{algorithm}[H]
\caption{\system: Search via Co-Evolving World Models}
\label{alg:system}
\begin{algorithmic}[1]
\State \textbf{Input:} Spec $\mathcal{T}$, Evaluator $\mathcal{E}$, Budget $B$, Stagnation Limit $K$
\State \textbf{Init:} $S \leftarrow \text{Init}(\mathcal{T})$
\While{$B > 0$}
  \State \textcolor[rgb]{0.6, 0.6, 0.6}{\Comment{1. Selection: Retrieve best action from frontier}}
  \State $a_t \leftarrow \operatorname*{arg\,max}_{a \in \mathcal{A}(S)} V(a \mid S)$
  \State $n \leftarrow 0$ \textcolor[rgb]{0.6, 0.6, 0.6}{\Comment{Stagnation counter}}
 \State $x_{\text{best}} \leftarrow \bot, \quad o_{\text{best}} \leftarrow \bot$
  
  \State \textcolor[rgb]{0.6, 0.6, 0.6}{\Comment{2. Instantiation: Local refinement loop}}
  \While{$B > 0$ \textbf{and} $n < K$}
    \State $x \sim \pi_{\text{code}}(\cdot \mid a_t)$
    \State $o \leftarrow \mathcal{E}(x)$; \;\; $B \leftarrow B-1$
    \If{$J(x) > J(x_{\text{best}})$}
      \State $x_{\text{best}} \leftarrow x$; \;\; $o_{\text{best}} \leftarrow o$
      \State $n \leftarrow 0$ \textcolor[rgb]{0.6, 0.6, 0.6}{\Comment{Reset counter on improvement}}
    \Else
      \State $n \leftarrow n+1$ \textcolor[rgb]{0.6, 0.6, 0.6}{\Comment{Increment on failure}}
    \EndIf
  \EndWhile
  
  \State \textcolor[rgb]{0.6, 0.6, 0.6}{\Comment{3. Evolution: Insert, Update, and Prune}}
  \State $S \leftarrow \text{P}_{\text{model}}(S, a_t, x_{\text{best}}, o_{\text{best}})$ 
\EndWhile
\State \textbf{return} Best found $x^\star$
\end{algorithmic}
\end{algorithm}
\end{minipage}
\vspace{-12pt}
\end{wrapfigure}

We build \system based on the concept of Search via Co-Evolving World Model. In our design, the LLM functions as an intrinsic world model that maintains and evolves a tree-structured state of the search process. The system operates by iteratively expanding, updating, and pruning an explicit search tree, decoupling high-level planning from low-level code generation. \cref{fig:system_overview} and \cref{alg:system} illustrate the complete workflow of \system.

\paragraph{Search State.}
The search state $S_t$ is maintained as an explicit search tree partitioned into \textsc{Closed} and \textsc{Open} nodes. \textsc{Closed} nodes (blue boxes in \cref{fig:system_overview}, e.g., $x_{12}$) represent visited actions where the local refinement process has concluded and the best-found programs are attached. In contrast, \textsc{Open} nodes (orange dashed boxes, e.g., $u_{13}$) form the frontier $\mathcal{A}(S_t)$ of pending actions waiting to be realized. Each \textsc{Open} node encapsulates a \emph{Proposed Optimization} tuple $(x_{\text{parent}}, \delta)$, linking a parent program to a specific natural language intent, and a \emph{Priority Score} $V \in [0,1]$. This score is dynamically updated by the world model (e.g., $u_{11}$ in the figure shows $V$ dropping from $0.9 \to 0.6$) to guide the selection of the next action.

\paragraph{Execution: Local Refinement.}
To isolate logical reasoning from implementation details, we treat program instantiation as a local refinement task. Upon selecting an action $a_t$ (Step 1 in \cref{fig:system_overview}), the system employs the LLM as a stochastic policy $\pi_{\text{code}}$. We repeatedly sample implementations $x \sim \pi_{\text{code}}(\cdot \mid a_t)$ and evaluate them ($o = \mathcal{E}(x)$) until a \emph{stagnation condition} is met ($K$ consecutive attempts without improvement). This ensures that valid actions are not discarded due to transient syntax errors or minor bugs.

\paragraph{Evolution: World Model Update.}
Upon concluding local refinement, the LLM analyzes the execution trajectory to update the state $S_t$ through three kinds of Tree Edit Operations (Step 3 in \cref{fig:system_overview}). It performs \emph{Insert} to propose new child nodes extending the current state (e.g., adding action $u_{13}$ with intent $\delta=$ ``fuse head'') and \emph{Update} to re-evaluate the priority of existing frontier nodes based on new evidence. Additionally, the model applies \emph{Prune} to identify and permanently remove infeasible or redundant branches (e.g., node $u_{10}$), thereby concentrating search resources on promising directions. Note that in the current version of \system, the world model evolution is merely performed by in-context learning of past observations.

\subsection{Case Study: MLA Paged Decode}
\label{sec:example-walkthrough}
\begin{figure*}[ht!] 
    \centering 
\includegraphics[width=0.95\linewidth]{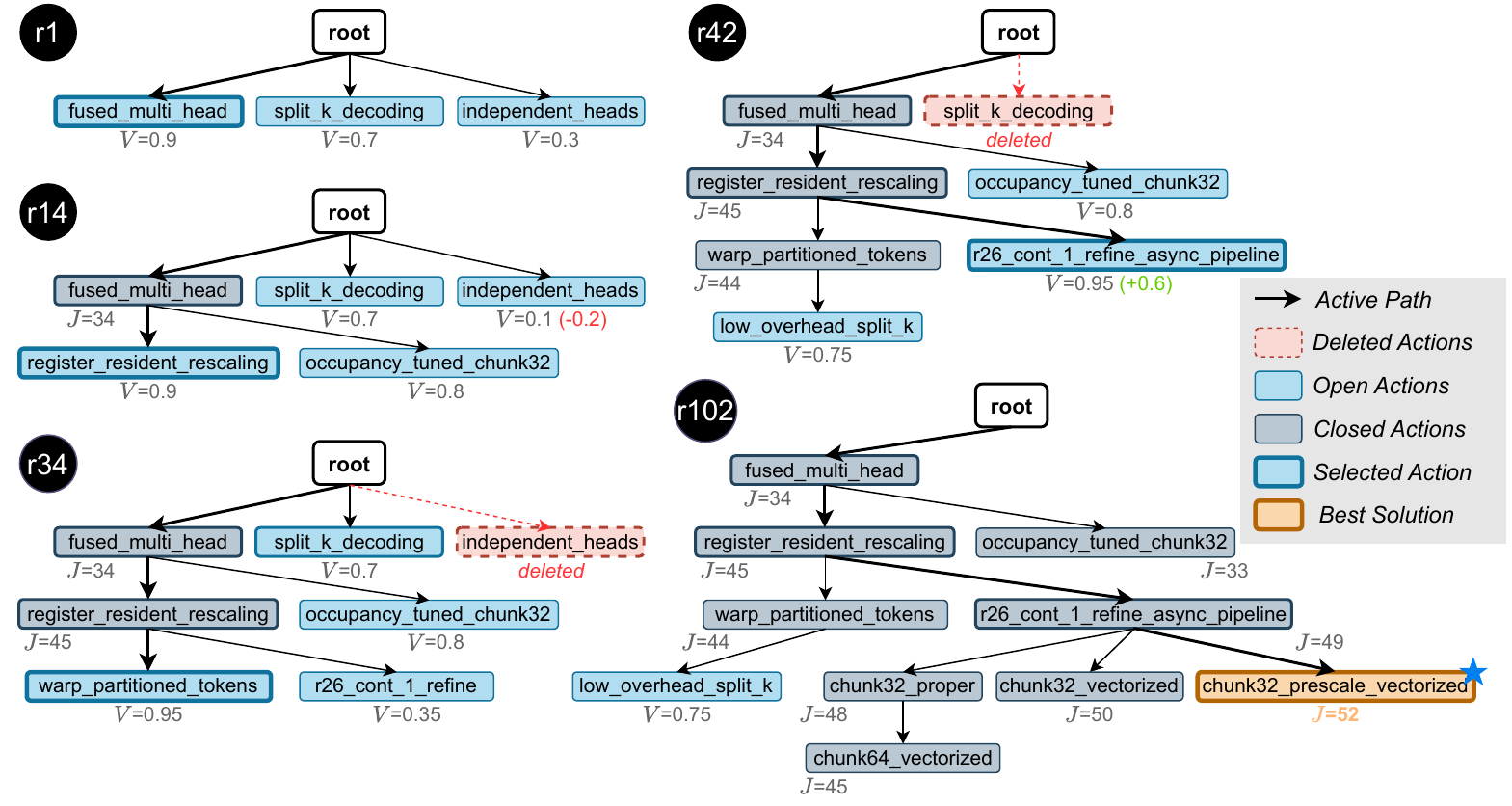} 
    \caption{\system Search Trace Visualization. It tracks the evolution of the Search State across search rounds on the MLA Paged Decode kernel (refer to \cref{sec:experiments} for setup details). A \textbf{round} corresponds to one candidate program evaluation. Nodes represent actions (dark=Closed, light=Open), annotated with their instantiated program performance (closed nodes) or priority scores (open nodes).
    The timeline highlights how the kernel is improved and how the LLM dynamically \emph{Inserts} new hypotheses, \emph{Updates} beliefs, and \emph{Prunes} less promising branches based on evolved understanding.} 
    \label{fig:vis} 
\end{figure*}

We illustrate the co-evolutionary search process using the MLA Paged Decode kernel (refer to \cref{tab:workloads} for kernel details). \cref{fig:vis} visualizes how the World Model co-evolves with the kernel implementation. We track the progress in rounds, where each round corresponds to a single invocation of the evaluator $\mathcal{E}(x)$ (i.e., one compiled and benchmarked kernel, consuming one unit of the budget). 

\paragraph{r1: Initialization and Hypothesis Selection.}
The search initializes $S_0$ with three high-level actions in the frontier: \emph{fused\_multi\_head}, \emph{split\_k\_decoding}, and \emph{independent\_heads}. The World Model assigns the highest value ($V$) to \emph{fused\_multi\_head}, hypothesizing that processing shared CKV heads together will reduce global memory traffic by $16\times$ compared to independent processing, and selects it for the first round of program instantiation.

\paragraph{r14-r34: Tree Evolution via Topological Edits.}
Following the local refinement of \emph{fused\_multi\_head} (score $J=34$), the model evolves the topology of $S_t$ to reflect this feedback. It first \emph{Inserts} refinement actions such as \emph{register\_resident\_rescaling} and \emph{occupancy\_tuned\_chunk32} to deepen the successful branch. Simultaneously, it \emph{Updates} the belief state by downgrading the sibling \emph{independent\_heads}, reasoning that the proven efficacy of head fusion renders independent processing less promising. By round 34, as evidence accumulates, the model permanently \emph{Prunes} the \emph{independent\_heads} branch, reallocating search resources entirely to the \emph{register\_resident} subtree.

\paragraph{r42-r102: Structural Insight and Experience Accumulation.}
At round 42, the World Model exhibits a structural shift in reasoning. It deletes the initial root-level \emph{split\_k} action but re-\emph{Inserts} a targeted variant, \emph{low\_overhead\_split\_k}, deep within the \emph{register\_resident} branch. This edit reflects a learned insight: split-K is ineffective as an isolated baseline but highly effective as a \emph{composable} optimization atop a strong fusion kernel. Notably, after \emph{chunk32\_vectorized} succeeds, the model proposes \emph{chunk32\_prescale\_vectorized} to apply \emph{sm\_scale} immediately upon loading $Q$. This specific refinement eventually yields the global optimum (star marker) at r102.

\paragraph{Takeaway.}
This trace highlights the efficiency of \system. By starting with high-level intent rather than raw code, the system avoids enumerating a massive sparse search space. It relies on local refinement to filter out transient coding noise and allows the World Model's understanding to \emph{co-evolve} with the kernel's optimization progress. Such evolution enables the system to prune dead ends and dynamically reposition strategies, guided by intrinsic reasoning over accumulated experience.

%% file: latex/exp.tex
\section{Experiments}
\label{sec:experiments}

\begin{table*}[ht!]
\centering
\caption{Representative kernels used for evaluation. Detailed configurations (e.g., head counts) are omitted for brevity. We provide example optimization challenges based on FlashInfer~\cite{flashinfer_deepwiki_mla} implementations.}
\label{tab:workloads}
\resizebox{\linewidth}{!}{%
\begin{tabular}{@{}l l l p{10.5cm}@{}}
\toprule
\textbf{Kernel} & \textbf{Arch.} & \textbf{Model Context} & \textbf{Characteristics \& Challenges} \\
\midrule
\textbf{MLA Paged Prefill} & Hopper & DeepSeek-V3~\cite{liu2024deepseek} & {\small Implements attention with paged KV cache and split \texttt{ckv} (compressed, no RoPE) and \texttt{kpe} (RoPE). Bandwidth-bound in long contexts; requires matrix absorption and Split-K optimization.} \\
\addlinespace
\textbf{MLA Paged Decode} & Hopper & DeepSeek-V3 & {\small Targets large dynamic batches. Latency-bound at low batch sizes. Uses Persistent Data Layout and specialized Hopper-instructions to minimize memory movement.} \\
\addlinespace
\textbf{GQA Paged Decode} & Hopper & Qwen3-A3B-30B~\cite{yang2025qwen3} & {\small Memory bound. Optimizations include fusing multiple query heads per KV head to reuse KV loads, and using vectorized/async paged-KV gather with pipelined global to shared memory prefetch. 
} \\
\addlinespace
\textbf{FP8 MoE} & Blackwell & DeepSeek-V3 & {\small Challenges include irregular data-dependent routing, load balancing, and managing FP8 packing/scaling overhead.} \\
\bottomrule
\end{tabular}%
}
\end{table*}

\subsection{Setup}
\label{sec:exp-setup}
Our main evaluation is conducted on FlashInfer~\cite{ye2025flashinfer} kernels, since these kernels are highly-optimized by experienced human engineers. For each target kernel, we run each system for a fixed budget of 120 iterations (we define one \textit{iteration} as one evaluation of a single candidate kernel on the benchmark workloads), and report the best-so-far score (i.e., $J(x)$ averaged over workloads in \cref{sec:problem_setup}) achieved at each iteration, using FlashInfer~\cite{ye2025flashinfer} kernels as reference SoTA.
We repeat each method three times and report the mean curve with a shaded min--max band to indicate the range of scores achieved across repeated experiments.

\paragraph{Implementation details.}
\system exposes a unified \texttt{Task} interface for plugging in optimization problems. Each task is defined by (i) a task specification and (ii) an evaluator. A task specification contains the PyTorch reference implementation, the optimization objective, and any task-specific instructions. The evaluator is responsible for compiling candidate implementations, validating functional correctness against the reference code, and measuring performance under a standardized benchmarking environment.

For FlashInfer kernels, \system integrates FlashInfer-Bench~\cite{xing2026flashinfer} as the task evaluator. All candidate implementations must be written in CUDA and pass functional correctness tests before receiving a non-zero score. We adopt the same compilation toolchain, correctness suite, and benchmark harness across all compared methods to ensure fairness. Experiments are conducted on NVIDIA H100 and B200 GPUs using CUDA 12.8, FlashInfer 0.5.3, and PyTorch 2.8.0.

\begin{figure*}[ht!]
    \centering
    \begin{subfigure}[t]{\linewidth}
        \centering
        \includegraphics[width=\linewidth]{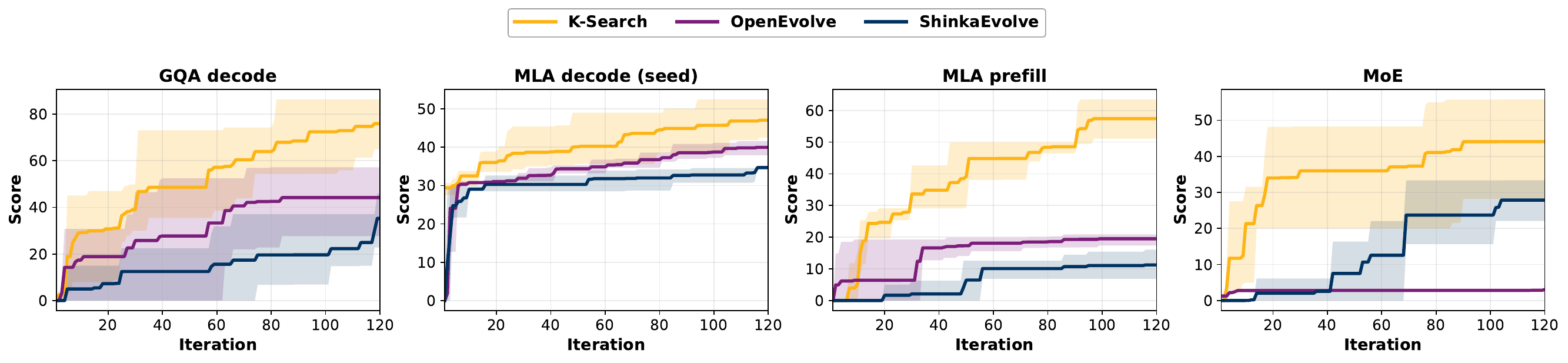}
        \caption{Search Process.}
        \label{fig:main_exp_a}
    \end{subfigure}
    \hfill
    \begin{subfigure}[t]{\linewidth}
        \centering
        \includegraphics[width=\linewidth]{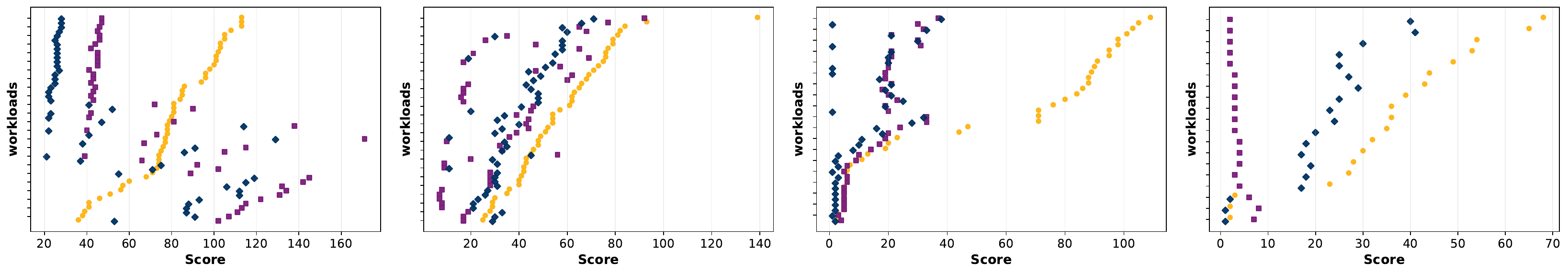}
        \caption{Best Kernel per-Workload Performance.}
        \label{fig:main_exp_b}
    \end{subfigure}
    \hfill
    \begin{subfigure}[t]{\linewidth}
        \centering
        \includegraphics[width=\linewidth]{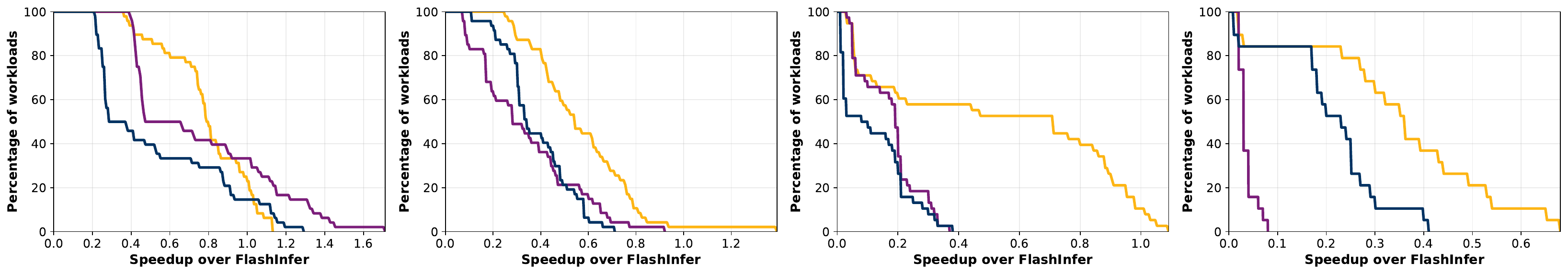}
        \caption{Best Kernel $\text{Fast}_p$ Plot.}
        \label{fig:main_exp_c}
    \end{subfigure}
    \caption{Main Results (3 runs each). (a) compares the kernels best-so-far scores generated by the three systems across 120 iterations. (b) provides a per-workload analysis for all compared systems. (c) shows the fraction of workloads for which the best kernel from each system achieves the specified speedup over the FlashInfer baseline. }
    \label{fig:main_exp}
\end{figure*}

\subsection{Baselines}
\label{sec:exp-baselines}
We evaluate 3 automated kernel-optimization methods: OpenEvolve~\cite{openevolve2025overview}, ShinkaEvolve~\cite{lange2025shinkaevolve}, and \system. We used \texttt{gemini-3-pro-preview} and the same initial program for each kernel. For ShinkaEvolve, we used \texttt{Qwen3-8B} as the embedding model. Across all experiments, we standardized the evaluation by using an identical set of input workloads for all methods on each kernel, and adhered to the default configurations for each baseline with the prompt template shown in~\cref{ssec:prompt-template}. \system sets budget value $B=120$ and stagnation value $K=7$. 

\subsection{Kernels}
\label{sec:exp-workloads}

We focus on four representative kernels (i.e., paged attention and MoE) from FlashInfer-Bench~\cite{xing2026flashinfer} that are widely used in modern LLM serving. We summarize these kernels in~\cref{tab:workloads}. Each kernel includes a fixed set of test traces used for correctness and benchmarking, captured in real traffic~\cite{xing2026flashinfer}.
We use identical traces for OpenEvolve, ShinkaEvolve, and \system to ensure a fair comparison. We provide an initial CUDA program for MLA decode kernel generation, as baselines struggle to write working kernels without an initial program. 

\subsection{Evaluation Results}
\label{sec:exp-results}


\textbf{Overall Performance.} \cref{fig:main_exp_a} compares the three systems across 120 iterations  
for GQA decode, MLA decode, MLA prefill, and MoE kernels (\cref{sec:exp-workloads}). 
For each method and kernel, we plot the scores over iterations across repeated runs. 
We find that \system significantly outperforms OpenEvolve and ShinkaEvolve. 
Across all kernels, \system achieves an overall average final score of 56.13, representing a \textbf{2.10 $\times$ improvement} over OpenEvolve (with score 26.68) and a \textbf{2.21$\times$ improvement} over ShinkaEvolve (with score 25.37). The performance gains vary across kernels. 
For the MoE kernel, \system achieves a final score of $44.1$, representing a $14.3\times$ improvement over OpenEvolve ($3.09$) and a $1.58\times$ improvement over ShinkaEvolve ($27.9$). 
Similarly, for MLA prefill, \system achieves a score of $57.4$ compared to OpenEvolve's $19.5$ and ShinkaEvolve's $11.3$, or $2.95\times$ and $5.10\times$ improvements, respectively. For GQA decode, \system reaches a score of $76.0$, outperforming OpenEvolve ($44.2$) and ShinkaEvolve ($27.7$) by $1.72\times$ and $2.74\times$. On the MLA decode, \system maintains a final score of $47.1$ versus $39.9$ and $34.7$, representing $18\%$ and $36\%$ improvements. 


\textbf{Best Kernel per-Workload Performance.} \cref{fig:main_exp_b} provides a per-workload analysis for all compared methods. Each dot represents the kernel's performance on a particular workload instance. Across $4$ kernel types and $152$ total workload traces, \system achieves higher performance than baselines on the vast majority of workloads. 
Interestingly, on some workloads for GQA decode, \system underperforms OpenEvolve and ShinkaEvolve, specifically those with small batch sizes: 16 of which have batch\_size=1, and 4 with batch\_size=16. \system does not underperform on workload with larger batch\_size. This occurs because \system's kernel employs a \emph{split-K} parallelism strategy that divides the key-value sequence across multiple thread blocks to maximize GPU utilization for large batches (described more in \cref{ssec:gqa_analysis}). While this approach excels when there is sufficient batch-level parallelism to amortize the coordination overhead, it introduces unnecessary synchronization costs for small batches.  
In contrast, OpenEvolve and ShinkaEvolve use a simpler single-block-per-batch design that processes the entire sequence within one thread block. 
While this simpler approach proves more efficient for batch\_size=1 (no coordination), it leads to bad performance with larger batch size. 


\textbf{Best Kernel $\text{Fast}_p$ Analysis.} \cref{fig:main_exp_c} shows the fraction of workloads for which the best kernel from each system achieves the specified speedup over the FlashInfer baseline. We observe that the generated kernels rarely exceed the expert-optimized FlashInfer kernel; however, \system significantly outperforms OpenEvolve and ShinkaEvolve. For GQA decode, \system attains a speedup $\geq 0.36$ on $100\%$ of workloads, compared to $50\%$ for ShinkaEvolve. At higher thresholds, the gap widens: at speedup $\geq 0.50$, \system succeeds on $87.5\%$ of workloads versus $50.0\%$ and $39.6\%$ for OpenEvolve and ShinkaEvolve, representing $1.75\times$ and $2.21\times$ improvements. 
For MLA prefill, at speedup $\geq 0.40$, \system reaches this threshold on $57.9\%$ of workloads, while none of the baseline solutions achieve speedup $\geq 0.40$.


\paragraph{Key observations.}
OpenEvolve and ShinkaEvolve both evolve over an archive of programs. 
The crucial difference from \system is that both directly search in program space and neither system \textit{explicitly tracks} which optimizations are likely to be correct or to improve performance. Consequently, ShinkaEvolve suffers from a low yield of correct programs: in our GQA logs, the vast majority of generations receive score zero (incorrect or failed programs). Search budget is heavily spent mostly on extending invalid or low-performing candidates. 
OpenEvolve 
similarly exhibits high per-iteration variance: Many iterations yield programs that underperform the currently best program. 
On harder tasks such as MoE, OpenEvolve’s search struggles to escape the low-accuracy regime: mean final score stays near 3 versus \system's 44.
We hypothesize that this advantage stems from \system's ability to maintain and co-evolve a persistent search state. By conditioning proposals on past attempts, the system generates more targeted hypotheses, significantly improving search efficiency.

\subsection{Kernel Analysis: FP8 MoE Kernel (Blackwell)}
\label{sec:moe_case_study}

We first discuss the FP8 MoE kernel: for each token, the top-$k$ experts are chosen from 256 candidate experts. The kernel then runs an ``up'' and ``gate'' projection, combined via SiLU, then a down-projection. All systems generate CUDA kernels for the same specification (DeepSeek-V3 style, top-8 routing, 32 local experts, hidden size 7168). We compare kernels generated by \system, OpenEvolve, and ShinkaEvolve.

\paragraph{Routing} 
Each token is assigned scores for 256 experts; the kernel then picks the top 8 experts. 
\system's kernel dedicates one GPU thread block per token (256 threads) and uses \emph{warp-level} cooperation: threads in a warp exchange values (\texttt{\_\_shfl\_down\_sync}) to find the 
the global top-8 experts. This keeps work parallel and avoids serialization. 

OpenEvolve uses a 
persistent kernel where each thread block runs in a loop: take the next “tile index” from a global counter (\textit{atomicAdd}), identify the (expert, token batch) it corresponds to, then perform computation for that tile. 
This incurs prohibitive overhead as it requires an atomic operation to process the next tile inside a while loop. In ShinkaEvolve, each thread loads all 256 scores for the experts and does plain \texttt{for}-loops to find the top experts, 
similarly leading to poor performance. 

\paragraph{Expert FFN computation} 
After routing, tokens must be processed by chosen experts. \system uses a simple pipeline: (1) routing, (2) sort-scatter that reorders tokens by expert to place them in contiguous memory, (3) computation (gate + up, then SiLU). \system uses tensor cores (with \texttt{WMMA}) on small $16\times16$ blocks and double-buffering so that loading the next block of data overlaps with computing the current one. \system's kernel also skips experts that receive zero tokens. As discussed, OpenEvolve adopts a single persistent kernel, which reduces kernel launches but needs more shared memory with lower GPU occupancy. 
ShinkaEvolve's kernel does not use tensor cores; each block performs a dot-product-style computation, thereby suffering from low performance. 

\subsection{Kernel Analysis: GQA Paged Decode (Hopper)}
\label{ssec:gqa_analysis}
We next discuss the \textit{GQA paged decode kernel}. In decode, each batch item contributes a single new query token, and the kernel attends over the keys and values already stored in the paged KV cache. 
All three systems---\system, OpenEvolve, and ShinkaEvolve---generate CUDA kernels for the same specification.

\paragraph{Parallelism over the sequence} 
In decode, the costly part is sweeping over the full key-value sequence. \system's kernel splits the sequence across multiple blocks: each block is assigned a contiguous chunk of keys and values, computes a partial attention result for that chunk, and writes it to a temporary buffer. When all chunks for a given (batch, key-value head) are done, one block detects that it is last (via a lightweight counter) and merges the partial results into the final output. Thus, for long sequences, many blocks work in parallel on different segments. OpenEvolve and ShinkaEvolve, however, use a single block per (batch, key-value head): that block loops over the entire key-value sequence itself. Therefore, for long sequences, they cannot exploit the same parallelism as \system. 

\paragraph{Overlapping memory and compute} 
\system loads keys and values in double-buffered chunks: while the current chunk is used for computation, the next chunk is being fetched. 
ShinkaEvolve does not double-buffer: it loads one chunk of keys and values, performs all attention work for that chunk, then loads the next. Memory load and compute are largely serialized, leading to lower performance. 

We defer discussion of other generated kernels to~\cref{ssec:other-generated-kernel-analysis}.


\subsection{GPUMODE TriMul}
\label{sec:gpumode}

\paragraph{Task Overview.}
\href{https://www.gpumode.com/home}{GPUMODE} is a public kernel optimization competition platform that hosts leaderboard-based challenges on real-world GPU workloads.
Participants submit Triton or CUDA implementations that must pass strict correctness checks before being evaluated under standardized benchmarking conditions.

The \emph{Triangle Multiplicative Update} (TriMul) is a core module in AlphaFold3~\cite{abramson2024accurate} and other protein structure prediction models.
It operates on a 4D pair representation $\mathbf{X} \in \mathbb{R}^{B \times N \times N \times C}$ and involves LayerNorm, five gated linear projections with optional masking, a pairwise contraction with $\mathcal{O}(N^3)$ complexity, and a final gated output projection.

\paragraph{Configuration.}
We run \system with $K{=}5$, reduced from $K{=}7$ used on FlashInfer CUDA tasks, as Triton's implementation is simpler than CUDA.
The search budget is 300 iterations: 150~steps with GPT-5.2, followed by 150~continuation steps with Gemini-3-Pro starting from the best solution found by GPT-5.2, without seed Triton programs provided.

\begin{wraptable}{r}{0.70\columnwidth}
\vspace{-10pt}
\centering
\small
\caption{Top GPUMODE TriMul submissions on NVIDIA H100. Latency is the geometric mean across a fixed set of benchmark cases.}
\label{tab:gpumode_h100}
\begin{tabular}{l l l r S}
\toprule
\textbf{Submission ID} & \textbf{Lang.} & \textbf{Model} & \textbf{Iter.} & {\textbf{Latency} ($\si{\micro\second}$, $\downarrow$)} \\
\midrule
shiyegao & CUDA & -- & -- & 1074 \\
Zeyu Shen & Triton & -- & -- & 1140 \\
TTT & Triton & GPT-OSS-20B w/ RL & 25{,}600 & 1161 \\
\midrule
\rowcolor{blue!8}\textbf{\system (Ours)} & Triton & GPT-5.2 + Gemini-3-Pro & 300 & \bfseries 1030 \\
\bottomrule
\end{tabular}
\vspace{-6pt}
\end{wraptable}

\paragraph{Results.}
We report leaderboard results from the official GPUMODE rankings\footnote{\url{https://www.gpumode.com/leaderboard/496?tab=rankings}} for competing submissions.
Since the submission period has closed, we evaluate our kernel locally using the official GPUMODE evaluator\footnote{\url{https://github.com/gpu-mode/reference-kernels/blob/main/problems/bioml/trimul/eval.py}} on NVIDIA H100 80\,GB HBM3 GPUs.
As shown in \cref{tab:gpumode_h100}, \system achieves state-of-the-art performance with a geometric-mean latency of \SI{1030}{\micro\second}, surpassing prior solutions, where the TTT submission is from a recent work TTT-Discover~\cite{yuksekgonul2026learning} that combines Reinforcement Learning (RL) with evolution methods.

%% file: latex/limitations.tex
\section{Conclusion}
\label{sec:conclusion}

In this work, we introduce \system, demonstrating that LLMs are not merely strong code generators but also possess latent planning capabilities that allow them to function as effective intrinsic world models. By replacing static search heuristics with a co-evolving world model, our framework enables the LLM to actively reason over the search space, distinguishing valid strategies from implementation noise. The empirical results, including a $2.1\times$ average improvement across diverse complicated kernels, up to $14.3\times$ gains on MoE over state-of-the-art evolutionary baselines, and SoTA performance on GPUMODE TriMul competition, validate this paradigm shift. These findings indicate that LLMs can serve as the core planning engine for complex optimization problems, moving beyond simple task implementation to autonomously driving the search strategy itself.

%% file: latex/acknowledgement.tex
\section*{Acknowledgment}
This work is supported by the kind compute support from Databricks, Amazon, Anyscale, and 
gifts from Google, Lambda, AMD, Mayfield, Accenture, Broadcom, Cisco, IBM, Intel, Intesa Sanpaolo, Lightspeed, Mibura, Microsoft, NVIDIA, Samsung SDS, and SAP. We also thank Dacheng Li, Mayank Mishra, Andy Yang, Parth Asawa, Fangzhou Zhao, and Tian Xia for helpful discussion and feedback.

%% file: latex/appendix.tex
\section{Appendix}

\subsection{Other generated kernels analysis}
\label{ssec:other-generated-kernel-analysis}

\paragraph{MLA Paged Prefill (Hopper):} 

In this kernel, each query is a 576-dimensional vector (512 CKV plus 64 ROPE). 
The kernel performs causal attention: it computes attention scores between queries and keys, applies softmax to get attention weights $P$, then forms the output as a weighted sum of values, $P \cdot V$. 
All three systems---\system, \emph{OpenEvolve}, and \emph{ShinkaEvolve}---generate CUDA kernels for the same specification (16 heads, causal masking, paged layout). 

\textit{Handling variable-length batches}: Each input batch has sequences of different lengths. A natural approach is to partition work into ``tiles'' of 16 query rows (one “row” of the attention computation for a single query vector). Because sequence boundaries do not align with tiles, a single tile can contain rows from more than one batch. \system handles this entirely on the GPU: each thread block is assigned a contiguous tile of 16 rows. When a 16-row tile straddles the end of one sequence and the start of the next, the thread block resolves the split on the fly. It uses the prefix-sum array of sequence boundaries to determine which subset of rows in the tile belongs to each sequence, then fetches the corresponding KV-cache range for that sequence and computes attention for that contiguous “segment” of rows. It then moves to the next segment within the same tile, repeating until the tile is done. 
\textit{OpenEvolve} and \emph{ShinkaEvolve} instead precompute a list of tile to batch mapping on the CPU and pass it to the GPU. However, that adds a separate CPU pass and extra memory to store the tile metadata. 

\textit{Score computation and softmax}: The expensive part of attention is computing attention scores (the equivalent of $Q K^\top$) and then softmax. 
\system keeps all threads in a block busy during this phase: they jointly compute the score matrix in small blocks, combine partial results in SRAM, then run the softmax per row. \textit{OpenEvolve} 
effectively restricts the score-and-softmax stage to one small group of threads (a single ``warp''); the rest of the block sits idle during that time, so a large fraction of the GPU's capacity in that block is unused. 


In summary, 
\system is faster because it (1) resolves batch boundaries on the GPU without a precomputed tile list or extra host-side pass; (2) keeps all thread groups busy during computation. 

\paragraph{MLA Paged Decode (Hopper):} 

Lastly, we discuss the \textit{MLA paged decode kernel}. In decode, each batch item has a single new query (one token). 
It computes attention scores $Q K^\top$ (over both \texttt{CKV} and \texttt{KPE}), applies softmax to get weights $P$, then forms the output as the weighted sum $P \cdot V$. All three systems---\system (best solution), \emph{OpenEvolve}, and \emph{ShinkaEvolve}---generate CUDA kernels for the same specification (16 heads, \texttt{CKV} dim 512, \texttt{KPE} dim 64). 

\textit{Sequence split and chunk size}. All three systems split the key-value sequence across multiple blocks: each block processes a contiguous chunk of the sequence, writes partial results to a temporary buffer, and a separate reduce step merges them into the final output and log-sum-exp. 
\system chooses when to split adaptively: for short sequences it uses a single block per (batch, head) and writes directly to the output, avoiding the reduce pass and extra memory. \textit{OpenEvolve} and \emph{ShinkaEvolve} use larger chunks (64 tokens) and fix a minimum number of splits, so they do not adapt well to short sequences. 

\textit{Avoiding shared-memory staging of the per-token query vectors (Q).} In MLA decode, the per-token query vectors (Q) are small (16 heads $\times$ 576 dims) but are reused across every chunk processed by a thread block. \system loads Q into register-resident fragments and reuses these fragments for all chunks in the block, without materializing the full Q in shared memory. In contrast, \textit{OpenEvolve} and \emph{ShinkaEvolve} stage the full Q matrix in shared memory at block entry. 
By keeping Q in registers, \system reduces per-block shared-memory pressure and achieves more speedups.

\textit{Overlapping memory and compute}. 
\system uses a deeper prefetch pipeline than standard double-buffering by loading \emph{two} chunks ahead: while the thread block is working on chunk $i$, it has already started loading chunk $i+1$ and \textit{issues} the load for chunk $i+2$. That keeps the pipeline fuller and hides more memory latency. \textit{OpenEvolve} and \emph{ShinkaEvolve} use standard double-buffering (load chunk $i+1$ while using chunk $i$). 
\system's deeper pipeline helps especially when the key-value sequence is long and many chunks are processed.


\system is faster because it 
(1) keeps the query in registers instead of SRAM. 
(2) adopts a deeper prefetch pipeline and overlaps memory and compute more aggressively by loading two chunks ahead, not just one; and (3) adapts the number of splits to sequence length. 

\subsection{Prompt template}
\label{ssec:prompt-template}

In this section, we show the prompt template we used for baselines for generation.

\begin{lstlisting}[style=icmlprompt,caption={Prompt used for code generation},label={lst:codegen-prompt}]
You are a code generator. Generate a CUDA kernel implementation optimized for {GPU} for the following specification.

Specification:
{specification}

Requirements:
- Write clean, efficient CUDA C++ code optimized for {GPU} architecture
- Use proper CUDA syntax and memory management optimized for {GPU}
- Implement the exact functionality described in the specification
- The reference code provides the mathematical specification but is unoptimized - your CUDA implementation should match its computational accuracy while delivering high performance
- Use the definition's tensor shapes, dtypes, and axes information to guide memory access patterns and optimization strategies
- Optimize for {GPU} GPU characteristics (memory hierarchy, compute units, etc.)
- For fixed axis values, optimize specifically for those constants rather than general cases
- You may use 3rd party libraries (cuBLAS, cuDNN, CUTLASS) when beneficial, but custom implementations often perform better for specialized kernels with known axis constraints

IMPORTANT: Generate code in XML format with exactly 3 files with these strict names:

<header_file name="kernel.h">
...
</header_file>

<cuda_file name="kernel.cu">
...
</cuda_file>

<cpp_file name="main.cpp">
...
</cpp_file>

Performance targets (lower is better):
{workloads}
\end{lstlisting}